# PUNJABI LANGUAGE INTERFACE TO DATABASE: A BRIEF REVIEW

**Preeti Verma,** Assistant Professor**,** Career Shapers, Amritsar (preet_asr156@yahoo.in)
**Suket Arora**, Research Scholar, Punjab Technical University, Kapoorthala (suket.arora@yahoo.com)
**Kamaljit Batra**, Assistant Professor, DAV College, Amritsar (kamaljit_batra@yahoo.com)

## ABSTRACT
*Unlike most user-computer interfaces, a natural language interface allows users to communicate fluently with a computer system with very little preparation. Databases are often hard to use in cooperating with the users because of their rigid interface. A good NLIDB allows a user to enter commands and ask questions in native language and then after interpreting respond to the user in native language. For a large number of applications requiring interaction between humans and the computer systems, it would be convenient to provide the end-user friendly interface. Punjabi language interface to database would proof fruitful to native people of Punjab, as it provides ease to them to use various e-governance applications like Punjab Sewa, Suwidha, Online Public Utility Forms, Online Grievance Cell, Land Records Management System, legacy matters, e-District, agriculture, etc. Punjabi is the mother tongue of more than 110 million people all around the world. According to available information, Punjabi ranks $10^{th}$ from top out of a total of 6,900 languages recognized internationally by the United Nations. This paper covers a brief overview of the Natural language interface to database, its different components, its advantages, disadvantages, approaches and techniques used. The paper ends with the work done on Punjabi language interface to database and future enhancements that can be done.*

## 1. INTRODUCTION

A natural language interface to a database (NLIDB) is a system that translates a sentence in natural language (e.g. Punjabi) into a database query. (Androutsopoulos et al.,1995[1]) .NLIDB bridge the gap between the languages used by humans and the computer systems as by using this, a person with no knowledge of database languages can easily and conveniently access the database. NLIDB can be considered as a classical problem in the field of natural language processing.[11] A person with no knowledge of Structured Query Language (SQL) may find himself or herself handicapped while extracting data from a Data Base Management System (DBMS) such as MS Access, Oracle and others. Therefore, Companies like Elfsoft(English Language Frontend Software) has developed SQL Tutor for people to interact with the database in simple English instead of learning the syntax of the query language. These products make the extraction from databases much easier. Therefore, the uses of the natural languages save time to learn the rigid syntax of query languages. NLIDB belong to the domain of information extraction as after grasping the input sentence, it extracts the requested information from data stores.

## 2. INTELLIGENT DATABASE SYSTEM (IDBS)

An intelligent database[13] is a full-text database that employs Artificial Intelligence (AI), interacting with users to ensure that returned items (hits) contain the most relevant information possible. Intelligent database (IDB) systems integrate the resources of both RDBMSs and KBSs to offer a natural way to deal with information, making it easy to store, access and apply.[11] "Intelligent database systems (IDBS) built from the integration of database (DB) technology with techniques developed in the field of artificial intelligence (AI)"[12]

### 2.1 An Ideal IDBS should [11]:
- Use knowledge and inference, making it easier to retrieve, view and make decisions with information
- Make information available to larger numbers of people because more people can now utilize the system due to its ease of use.
- Improve the decision making process involved in using information.
- Interrelated information extract from different sources using different media so that the information is more easily utilized by the user.

- Provide high-level intelligent tools that provide new insights into the contents of the database by extracting knowledge from data.

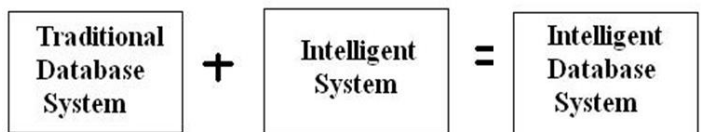

Fig. 1 – IDBS

## 3 TYPES OF INTERFACES

A user interface is the view of a database interface that is seen by the user.
- Form-based Interfaces
- Text-based Interfaces
- GIS Interface

### 3.1 Form-based Interfaces:
This interface consists of forms which are adapted to the user. User can fill in all of the fields and make new entries to the database. They are easy to use and have the advantage that the user does not need special knowledge about database languages like SQL. Eg: SBB Travel Online website. But some operations might be restricted by the application.

### 3.2 Text-based Interfaces:
Using this interface, the users communicate with the DBMS directly in the query language via an input/output window. Text-based interfaces are very powerful tools and allow a comprehensive interaction with a DBMS. However, the use of those is based on active knowledge of the respective database language. For Eg: SQL.

### 3.3 GIS Interface:
A GIS user interface often integrates features of a database interface. The database interaction takes place through the combination of different interfaces:
- Graphical interaction via a selection on the map
- Combination of form-based and text-based interaction (e.g. special Query-Wizards for the easier creation of database queries)[2]

## 4. COMPONENTS OF NLIDB

The problem of natural language access to a database is divided into two sub-components:
- Linguistic component
- Database component

### 4.1 Linguistic Component:
This component translates the natural language input sentence into a formal query and then after a database search generates a natural language response.

### 4.2 Database Component:
This component performs traditional Database Management functions. Natural language database systems make use of syntactic knowledge and knowledge about the actual database in order to properly relate natural language input to the structure and contents of that database. Questions entered in natural language translated into a formal query. This query is then processed by the database management system and after processing, the result is passed back to the natural language component where generation routines produce a surface language version of the response.

## 5. ADVANTAGES AND DISADVANTAGES OF NLIDB

This section covers some advantages and disadvantages of NLIDBS, comparing them to different types of interfaces.

### 5.1 ADVANTAGES OF NLIDB:
- **No need to learn Artificial Communication language**
- **Less training required**
- **Easy to handle negation and quantification type of questions**
- **Easy to Use for Multiple Database Tables**
- **Easy and efficient retrieval**
- **Tolerances to minor grammatical errors**
.
### 5.2 DISADVANTAGES OF NLIDB:
- **Deals with limited set of natural language**
- **Linguistic vs. conceptual failures**
- **Users assume intelligence**
- **Inappropriate Medium**
- **Tedious configuration**

## 6. APPROACHES FOR DEVELOPMENT OF NLIDB SYSTEM

## 6.1 Symbolic Approach (Rule Based Approach)

Knowledge about language is explicitly encoded in rules or other forms of representation. Language is analyzed at various levels to obtain information. On this obtained information certain rules are applied to achieve linguistic functionality. Rules are formed for every level of linguistic analysis. It tries to capture the meaning of the language based on these rules.

## 6.2 Empirical Approach (Corpus Based Approach)

A corpus is collections of machine readable text. Corpora are primarily used as a source of information about language. This approach is based on statistical analysis as well as other data driven analysis, of raw data which is in the form of text corpora. The empirical or corpus based methods automate the acquisition of much of the complex knowledge required for NLP by training on suitably annotated natural language corpora, e.g. tree-banks of parsed sentences. Various statistical techniques like hidden Markov models, probabilistic context free grammars etc are employed by most of the empirical NLP methods.

## 6.3 Connectionist Approach (Using Artificial Neural Network):

Since human language capabilities are based on neural network in the brain, Artificial Neural Network provides on essential starting point for modeling language processing. Instead of symbols, the approach is based on distributed representations that correspond to statistical regularities in language. However, there has been relatively little recent language research using sub-symbolic learning, although some recent systems have successfully employed decision trees transformation rules and other symbolic methods. SHRUTI system developed by John Carter Wendelken is a neutrally inspired system for event modeling and temporal processing at a connectionist level.

## 7. ARCHITECTURE OF NLIDB SYSTEM[5]

### 7.1 Pattern Matching Systems

In pattern matching systems, some rules and patters are initially given. These patterns and rules are fixed. The rules states that if an input word or sentence is matched with the given pattern, the action has been taken. The main advantage of the pattern-matching approach is its simplicity: no elaborate parsing and interpretation modules are needed, and the system is easy to implement. Eg: SANVY system.[1]

### 7.2 Syntax-Based Systems

In syntax-based systems, the user's question is parsed and the resulting parse tree is directly mapped to an expression in some database query language. Syntax-based systems use a grammar that describes the possible syntactic structures of the user's questions. The main advantage of using syntax based approaches is that they provide detailed information about the structure of a sentence. Eg: LUNAR system.[1]

### 7.3 Semantic Grammar Systems

The basic idea of a this system is to simplify the parse tree as much as possible, by removing unnecessary nodes or combining some nodes together. Less ambiguous compared to the syntax based approach. The main drawback of this approach is it requires some prior- knowledge of the elements in the domain, therefore making it difficult to port to other domains. A parse tree in this system has specific structures and unique node labels, which could hardly be useful for other applications. Eg: LADDER system.[6]

### 7.4 Intermediate Representation Languages

In this approach, the system can be divided into two parts:
- One part starts from a sentence up to the generation of a logical query.
- Other part starts from a logical query until the generation of a database query.

The intermediate logical query expresses the meaning of the user's question in terms of high level world concepts, which are independent of the database structure. The logical query is then translated to an expression in the database's query language, and evaluated against the database.(SQL)

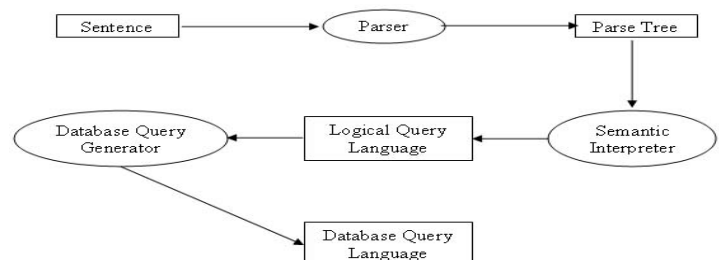

**Figure 2 - Intermediate Representation Language Architecture**

Example of intermediate representation language architecture is Masque/sql[1].

## 8. EXISTING PUNJABI LANGUAGE INTERFACE TO DATABASE REVIEW:

In India there is research going on in this topic in many Indian languages like Hindi, Punjabi, Malayalam, Bengali etc. In Punjabi language a system is developed at Thapar University, Patiala.[3] The system used the agriculture domain, having Farmer, Crop and Sale tables. It accepts query in Punjabi language that is translated into SQL query, by mapping the Punjabi language words, with their corresponding English words with the help of database maintained. Then, the query is executed.

### 8.1 Architecture

Architecture of this system is divided into three phases:
- Tokenizing the query
- Mapping and formation of SQL query
- Execution of query.

### 8.2 Queries Handled

This system has been tested for 3 types of queries:
- Queries for selection of all the columns
- Queries for selection of certain columns
- Queries for selection of certain rows from certain columns

## CONCLUSION

The existing Punjabi language interface to database system[3] has limited scope. It is domain dependent and uses only few tables and simple selection queries. In future, the existing Punjabi language interface to database system developed by Thapar University, Patiala[3] can be extended to perform the following tasks:
- It can execute Queries that involves joining of tables and handle various clauses
- It can be made domain independent
- It can be extended to handle more complex queries
- Development of an interface for administrator
- This existing system accepts queries only in a particular format. So to make the system to accept all the queries, we can use Punjabi parser

## REFERENCES

[1]. Androutsopoulos, G.D. Ritchie, and P. Thanisch, Natural Language Interfaces to Databases – An Introduction, Journal of Natural Language Engineering 1 Part 1 (1995), 29–81.
[2] GITTA - Geographic Information Technology TrainingAlliance(http://www.gitta.info - Version from:12.7.2007)
[3] Amandeep kaur "Punjabi Language Interface to databases", ME Thesis,Thapar University, june 2010.
[4]. Himani Jain, Parteek Bhatia "Hindi Punjabi Language Interface to databases",Journal of Global Research in Computer Science, Volume 2, No. 4, April 2011 Part 1 (1995), 29–81.
[5] B.Sujatha, Dr.S.Viswanadha Raju, Humera Shaziya "A Study of the Various Architectures for Natural Language Interface to DBs", (IJCSN) Volume 1, Issue 4, August 2012
[6] Mrs. Neelu Nihalani , Dr. Sanjay Silakari, Dr. Mahesh Motwani "Natural language Interface for Database: A Brief review", IJCSI, Vol. 8, Issue 2, March 2011
[7] Mrs. Neelu Nihalani , Dr. Sanjay Silakari, Dr. Mahesh Motwani " An Intelligent Interface for relational databases", IJSSST, Vol. 11, No. 1.
[8] Arup Panja, Rami Reddy Nandi Reddy and Sivaji Bandyopadhyay "A Natural Language Interface to a Database System", *Journal of Intelligent Systems, Vol. 16, No. 4, 2007.*
*[9]* MOLOY DEY "NATURAL LANGUAGE BASED DATABASE QUERY",Master of Technology in IT, JADAVPUR UNIVERSITY KOLKATA,2010.
[10] Mrs. Neelu Nihalani , Dr. Sanjay Silakari, Dr. Mahesh Motwani " Design of Intelligent layer for flexible querying in databases", /International Journal on Computer Science and Engineering, Vol.1(2), 2009
[11].Kamran Parsaye, Mark Chignell, Setrag Khoshafian and Harry Wong, "Intelligent databases-object-oriented, deductive hypermedia technologies", New York, John Wiley& Sons, 1989.
[12] Bertino, B. Catania, G.P. Zarri, "Intelligent database systems", Reading, Addsion Wesley Professional, 2001.